\documentclass{article}
\PassOptionsToPackage{numbers, compress}{natbib}
\usepackage[preprint]{neurips_data_2024}

\usepackage[utf8]{inputenc} 
\usepackage[T1]{fontenc}    
\usepackage{hyperref}       
\usepackage{url}            
\usepackage{booktabs}       
\usepackage{amsfonts}       
\usepackage{nicefrac}       
\usepackage{microtype}      
\usepackage{xcolor}         
\usepackage{amsmath}

\usepackage{pifont}
\usepackage{graphicx}
\usepackage{algorithm}
\usepackage{algorithmic}

\title{Open-Event Procedure Planning \\in Instructional Videos}

\author{Yilu Wu$^1$,
Hanlin Wang$^1$,
Jing Wang$^1$,
Limin Wang$^{1,2,}$\thanks{Corresponding author (lmwang@nju.edu.cn).}\\
$^1$State Key Laboratory for Novel Software Technology, Nanjing University \\
$^2$Shanghai AI Lab \\
\textbf{\url{https://github.com/MCG-NJU/OEPP}}\\
}

\begin{document}

\maketitle

\begin{abstract}
 Given the current visual observations, the traditional procedure planning task in instructional videos requires a model to generate goal-directed plans within a given action space. All previous methods for this task conduct training and inference under the same action space, and they can only plan for pre-defined events in the training set. We argue this setting is not applicable for human assistance in real lives and aim to propose a more general and practical planning paradigm. Specifically, in this paper, we introduce a new task named Open-event Procedure Planning (OEPP), which extends the traditional procedure planning to the open-event setting. OEPP aims to verify whether a planner can transfer the learned knowledge to similar events that have not been seen during training. We rebuild a new benchmark of OpenEvent for this task based on existing datasets and divide the events involved into base and novel parts. During the data collection process, we carefully ensure the transfer ability of procedural knowledge for base and novel events by evaluating the similarity between the descriptions of different event steps with multiple stages. Based on the collected data, we further propose a simple and general framework specifically designed for OEPP, and conduct extensive study with various baseline methods, providing a detailed and insightful analysis on the results for this task.
\end{abstract}

\section{Introduction}
\label{sec:intro}

Humans can learn procedural knowledge from instructional videos and figure out what actions should be performed to achieve their desired goals. This ability is crucial for the next-generation AI system as such a model can analyze complex human behaviors and help people with goal-directed problems like cooking. Recent works~\citep{DDN,GAIL,plate,p3iv,PDPP,E3P,maskpdpp,skipplan,schema} have shown great promise for the procedure planning problem in instructional videos, which requires a model to generate proper action sequences to transform from the given start state to goal state. In this task, all actions that make up the planning sequences are selected from a given action space provided by the instructional video dataset, both for training and inference.

\begin{figure}[t]
  \centering
   \includegraphics[width=0.65\linewidth]{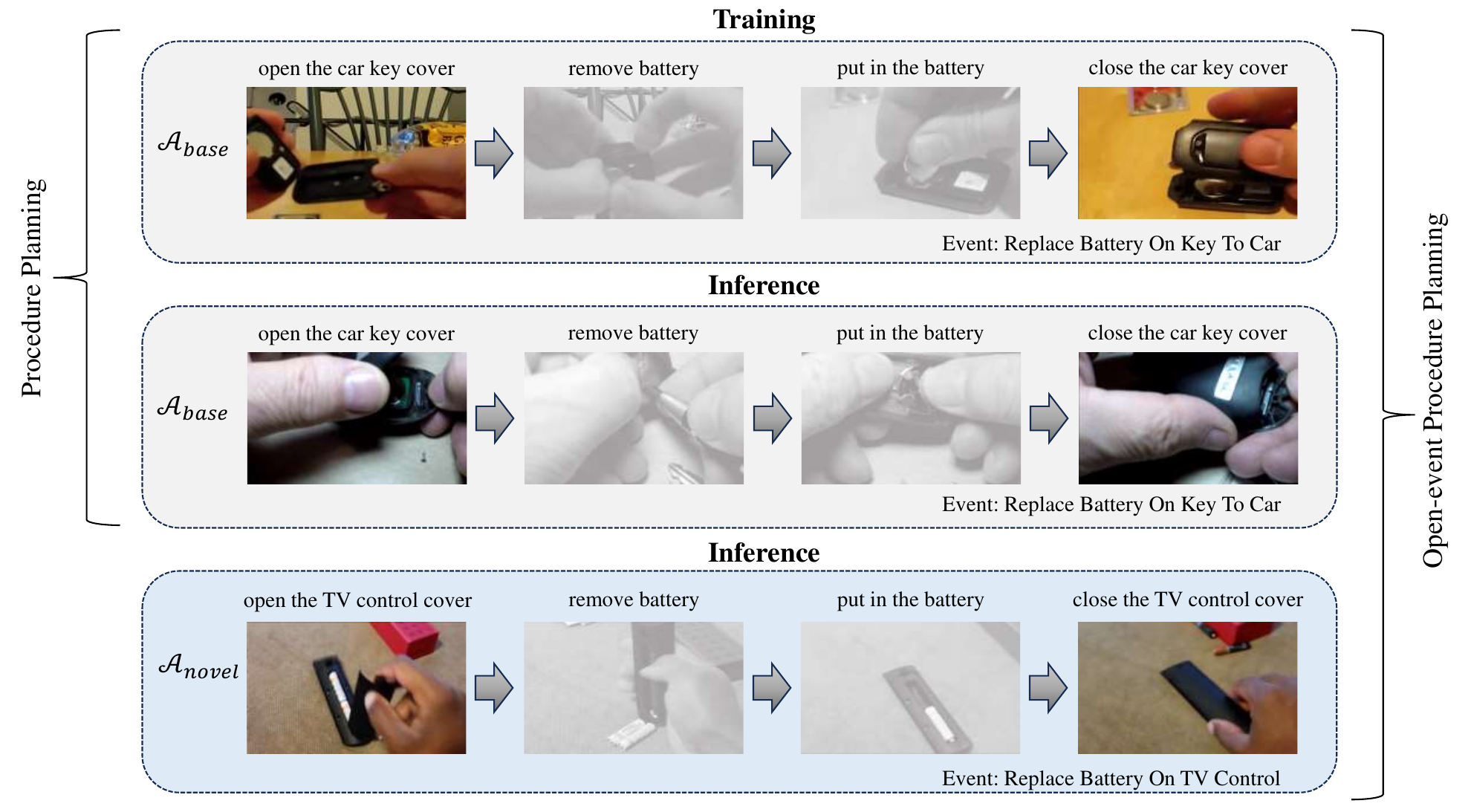}
   \caption{Illustration of procedure planning and open-event procedure planning. Procedure planning train and infer model under the same action space, while open-event procedure planning conducts inference under both base (\textit{Replace Battery On Key To Car}) and novel (\textit{Replace Battery On TV control}) action spaces.}
   \label{fig:intro-sample}
   \vspace{-1.2em}
\end{figure}

Given that models for the initial procedure planning problem are trained and inferred with the same action space, actually these models do not need to understand the meaning of actions performed at each step, but only needs to imitate the process in training videos. However, in addition to completing tasks through instructions, it is more important that humans can transfer their learned knowledge to similar events they have never seen before. As shown in Figure~\ref{fig:intro-sample}, for the new event \textit{Replace Battery On TV Control} whose steps are very similar to \textit{Replace Battery On Key To Car} which has seen before, we can roughly deduce the entire sequence of actions: \textit{open the cover first} → \textit{remove the old battery} → \textit{put new battery in} → \textit{close the cover}. This transfer ability requires people to fully understand the meaning of each action and master the procedural knowledge learned from seen videos, thus helps people better handle real-life situations. In this sense, methods for procedure planning which are all based on a close-set assumption can only identify and plan under events and actions presented in the training set. This greatly limits the application scope of procedure planning since covering all real-life events in the training set is really hard.

Recently, visual language models (VLMs)~\cite{CILP,ALIGN,MIL-NCE,videoclip} pre-trained on large-scale vision-text pairs have shown their remarkable zero-shot performance recently. These VLMs align vision and language features into the same space, fulfilling the gap between visual and language data. Many open vocabulary~\cite{ovdetr,ovtrack,IVF,pointclip,actionclip} approaches thus eliminate the distinction between close-set and open-set by exploiting the aligned features learned by VLMs. Inspired by these methods, we extend the procedure planning task from close-set to open-set, with an aim to develop planning models that can truly understand action meanings and transfer procedural knowledge to new tasks like human. Compared with the initial procedure planning, such a planning setting is more general and practical.

In this paper, we propose a new task named Open-event Procedure Planning (OEPP). As shown in Figure~\ref{fig:intro-sample}, we extend procedure planning to the open-event setting which requires the model to plan action sequences for novel events unseen during training. Specifically, we train model under the base action space and conduct inference under base and novel action spaces to evaluate their transfer ability. The base and novel action spaces are provided during training and inference, respectively. It is worth noting that we still provide an action space for novel events. This is because if we directly require the model to plan without an action space, the description and variety of actions generated by the model may differ significantly from the manually action annotation results in the dataset, making it hard to evaluate the correctness of the generated results. Moreover, since our main motivation is to evaluate whether the model can transfer learned procedure knowledge to events and actions it has not seen before rather than action generation, we still provide novel action space for better evaluation.

Another point worth noting is that we should ensure the procedural knowledge of base events can transfer to novel events, indicating a key aspect of ``transfer ability''. For example, we can learn \textit{How to replace battery on TV control} from \textit{How to replace battery on key to car}, but can not learn \textit{How to make a cake} from it. However, the existing instructional video datasets~\cite{MPII,breakfast,50salads,Youcook,Youcook2,crosstask,COIN,NIV} do not take this into account, thus not suitable for our new setting. Simply dividing base and novel events with existing datasets will create a significant gap between the training and test sets, making the transfer hard and meaningless. Therefore, we rebuild a new benchmark named OpenEvent by combining two large instructional video datasets, COIN~\cite{COIN} and CrossTask~\cite{crosstask}, to evaluate the open-event procedure planning problem. We utilize an approach based on text similarity and human verify combination to create the dataset, which involves four main stages: text similarity clustering, human verification, action descriptions refinement, and dataset split. Details will be shown later.

We further propose a simple and general framework specifically designed for open-event procedure planning. Specifically, we transform this problem into a visual-text matching problem, and apply several procedure planners, such as simple MLP-based method, Transformer-based method and modified version of PDPP~\cite{PDPP} to generate $T$ embeddings as the action plan. Here $T$ denotes the length of planning horizon. Then the output will be used to calculate the similarity with text features in the given action space. We match the output with correct actions during training phase and select action with the highest similarity at each step to get the final action sequence for inference.

Our \textbf{contributions} are summarized in three aspects: (i) We propose a new task called Open-event Procedure Planning (OEPP), which extends procedure planning to an open-event setting. (ii) We rebuild a new benchmark termed as OpenEvent based on COIN~\cite{COIN} and CrossTask~\cite{crosstask} for OEPP. (iii) We propose a simple and general framework specifically designed for OEPP, and conduct extensive study with various baseline methods, providing a detailed and insightful analysis on the results for this new task. We hope our work can inspire more works on OEPP, which is more practical for real-life applications. Our code and data: https://github.com/FOXamber/OEPP.

\section{Related work}
\label{sec:related}

\noindent\textbf{Procedure planning.} The initial planning task~\cite{finn2017deep,finn2016deep} plans the actions in a very simple environment, such as stacking blocks on a table. Later, Chang et al.~\cite{DDN} propose to migrate procedure planning to more complex real life, so procedure planning is naturally scaled to instructional videos. Previous methods can be divided according to the supervision method: (i) one-hot label and visual information supervision~\cite{DDN,GAIL,plate}, these methods generate both action sequences and visual states for intermediate steps during training. (ii) one-hot label and text supervision~\cite{p3iv,schema}, which use a non-autoregressive transformer-based~\cite{attention} architecture. (iii) one-hot label supervision~\cite{PDPP,maskpdpp,skipplan}, these approaches treat this problem as a distribution fitting problem using diffusion model~\cite{ddpm,iddpm} or transformer. (iv) text information supervision~\cite{E3P}, which plans out actions based on both the states and predicted events. However, all the above methods are based on the close-set assumption, that is, these methods cannot plan action sequences for events that have not appeared in the training set.

\noindent\textbf{Open vocabulary learning.} For visual scene understanding, most approaches focus on the close-set assumption, meaning that the model can only identify pre-defined categories that are present in the training set. Recently, due to the rapid progress of vision language pre-training models (VLMs)~\cite{CILP,ALIGN,MIL-NCE,videoclip}, open vocabulary settings were proposed including a wide range of computer vision tasks, object detection~\cite{ovdetr,regionclip}, segmentation~\cite{maskclip}, video understanding~\cite{actionclip,IVF}, and 3D scene understanding~\cite{pointclip}. Following open vocabulary learning tasks, we propose the concept of open-event procedure planning. In contrast to traditional open vocabulary, we aim to open the event category rather than the action category. Due to this difference, our base and novel categories are not entirely independent and there is a small overlap. 

\noindent\textbf{Instructional videos.} Instructions can take various forms such as text, voice, and video. Video instructions are more intuitive and easier to comprehend, which has led to the creation of many instructional video datasets in recent years~\cite{breakfast,50salads,Youcook, Youcook2, crosstask,COIN,NIV,howto100m}. In particular, HowTo100M~\cite{howto100m} provides a large amount of training data for video-text representation learning. We summarize existing instructional video datasets and find that early datasets primarily focused on kitchen scenes, resulting in a relatively narrow domain. However, with the proposal of CrossTask~\cite{crosstask} and COIN~\cite{COIN}, more and more domains and events are considered, which also provides huge basic data for our dataset.

\section{Open-event procedure planning}

In this section, we provide a detailed overview of our Open-event Procedure Planning (OEPP). We first present the definition of OEPP task, then we elaborate on the data collection process of our OpenEvent benchmark. Finally, we introduce the evaluation metrics we applied for this task.

\subsection{Task definition}
In general, given the start and end observations, $o_{start}$ and $o_{end}$, the model need to plan an action sequence $\pi$ under an action space $\mathcal{A}$ to transfer from $o_{start}$ to $o_{end}$: $\pi =\left \{ a_{1} ,a _{2},...,a _{T} \right \} ,a _{i}\in \mathcal{A}.$ Here $T$ defines the number of actions we need to plan. The initial procedure planning train and infer model under the same action space $\mathcal{A}_{base}$, which is composed of subspaces of $E_{b}$ base events, $\mathcal{A}_{base} = \left \{ \mathcal{A}_{1}\cup \mathcal{A}_{2} \cup ...\cup \mathcal{A}_{E_{b}} \right \}.$ To extend to the open-event setting, we also evaluate the model under the novel action space $\mathcal{A}_{novel}$, which is composed of subspaces of  $E_{n}$ novel events, $\mathcal{A}_{novel} = \left \{ \mathcal{A}_{E_{b}+1}\cup \mathcal{A}_{E_{b}+2} \cup ...\cup \mathcal{A}_{E_{b}+E_{n}} \right \}.$

\begin{table}[t]
\caption{Comparisons of existing instructional video datasets, ``Hierarchical'' refers to whether the dataset contains a hierarchical structure, such as ``Domain-Event-Action''. ``Transferable'' refers to whether the dataset considers the transfer ability between events.}
\centering
\resizebox{1\linewidth}{!}{
\begin{tabular}{rcccccccc}
\hline
\textbf{Dataset} & \textbf{Domains} & \textbf{Events} & \textbf{Actions} & \textbf{Videos} & \textbf{Segments} & \textbf{Annotation} & \textbf{Hierarchical} & \textbf{Transferable} \\ \hline
MPII \cite{MPII}             & 1       & 14              & 65               & 44              & 5609              &  \checkmark                  &  \ding{55}                    &   \ding{55}             \\
YouCook \cite{Youcook}          & 1       & -               & -                & 88              & -                 & \ding{55}                   & \ding{55}                    & \ding{55}               \\
50salads \cite{50salads}         & 1       & 1               & 17               & 50              & 966               &   \checkmark                    &  \ding{55}                      &  \ding{55}              \\
Breakfast \cite{breakfast} &  1       & 10              & 48               & 1989            & 11267             &   \checkmark                   &  \ding{55}                      & \ding{55}               \\
NIV \cite{NIV}             & 5               & 5               & 48               & 150             & -                  & \checkmark                   &  \ding{55}                      &\ding{55}                \\
YouCook2 \cite{Youcook2}        & 1       & 89              &  -                & 2000            & 13829             &  \checkmark                   & \ding{55}                      &  \ding{55}              \\
EPIC-KITCHENS \cite{EPIC-KITCHENS}   & 1       & -                & -                 & 432             & 39596             &  \checkmark                   &  \ding{55}                      & \ding{55}               \\
CrossTask \cite{crosstask}       & 4                & 18              & 105              & 2763            & -                  &\checkmark                    &   \ding{55}                     & \ding{55}               \\
COIN \cite{COIN}             & 12               & 180             & 778              & 11827           & 46354             &\checkmark                    &  \checkmark                     & \ding{55}               \\
HowTo100M \cite{howto100m}       & 12                &23611                 & -                 & 1.22M           & 136.6M            & \ding{55}                   &\checkmark                       & \ding{55}               \\
Assembly101 \cite{assembly101}     & 1                &  4               & 1456                 &4321                 & 1M                  &\checkmark                     &\ding{55}                       &\ding{55}                 \\ \hline
Ours      & 8                &  43               & 161                 &2771                 & 12210                  &\checkmark                     & \checkmark                      &\checkmark                \\\hline

\end{tabular}}
\label{tab:datasets}
\end{table}

\subsection{Benchmark: data}
\label{sec:data}

In this part, we introduce the reconstruction process of the dataset in detail. Considering the diversity of events, domains and the consistency of annotation, we use COIN~\cite{COIN} and CrossTask~\cite{crosstask} as our raw data. COIN includes 180 events and CrossTask includes 18 events, involving a total of 12 different domains. We remove duplicate tasks and tasks with less than three actions, such as \textit{Change A Tire} in CrossTask which is duplicated with the \textit{Change Car Tire} in COIN, and \textit{Prepare Canvas} in COIN which only contains two actions. Additionally, we remove events that do not contain procedural knowledge, such as \textit{Use Earplugs}, \textit{Put On Hair Extensions} and \textit{Practice Karate} in COIN. After preliminary screening, there are a total of 189 events that can be used in subsequent stages. Then we conduct the dataset construction process with the following four stages.

\noindent\textbf{Stage one: text similarity clustering.} Human verification of 189 events can be time-consuming and subjective. To address this issue, we propose using text similarity as an initial clustering method. We start by rewriting each action into a sentence that includes the event and the guiding action sequence provided by the dataset. For example, the sentence of \textit{Draw Blood}: ``Draw Blood: 1.tie the tourniquet 2.disinfect 3.collect blood 4.pull out the needle and press with cotton.'' Then we use the existing sentence encoder Sentence-Bert~\cite{sentence-bert} to encode each sentence into a vector. Finally, we calculate the cosine similarity to group events with high similarity into the same cluster. As shown in the algorithm in supplementary material, we set the threshold $\theta$ to 0.6 and group all event sentences into different clusters. For each cluster, the events within it have similar steps. Clusters with only one event means that there are no other events similar to it. In such cases, we temporarily remove these events.

\noindent\textbf{Stage two: human verification.} Comparing the text similarity of two sentences can help identify whether two events can be transferred, but cannot serve as the final judgment. Based on clusters obtained in the first stage, we further verified whether the events in the clusters can truly be transferred. With human verification, we delete some unreasonable events such as \textit{Make Flower Crown} and \textit{Make Flower Press}. The text similarity between these two events is high since the word ``flower'' appears both in the two event sentences, while actually these events cannot learn procedural knowledge from each other. Additionally, we add some new clusters, such as \textit{Replace Graphics Card} and \textit{Replace Memory Chip}. Ten annotators are involved in this validation process to assess whether tasks within the same cluster are transferable, and results considered transferable by over half of the annotators are kept. After this stage, we get 14 clusters with a total of 43 events and 161 actions.

\noindent\textbf{Stage three: action description refinement.} We notice that some action descriptions in events are not reasonable. For example, task \textit{Replace Battery On TV Control} in COIN contains action \textit{put battery in}, while task \textit{Replace Battery On Key To Car} contains action \textit{put in the battery}. The two actions are actually the same action but have different descriptions, which may interfere with our evaluation. We thus conduct a refinement on all action descriptions to ensure these descriptions accurate and consistent. The final cleaned results are in the supplementary materials.

\noindent\textbf{Stage four: dataset split.} In order to ensure that all novel events can be transferred from one or more base events, we select one event from each cluster as novel event, and all remaining events as base events. For dataset split, we select 80\% samples of base events for training and 20\% for testing. Samples of novel events are used for testing only. We then select 20\% samples from the training set as the validation set for model selection.

\begin{figure}[t]
  \centering
   \includegraphics[width=0.9\linewidth]{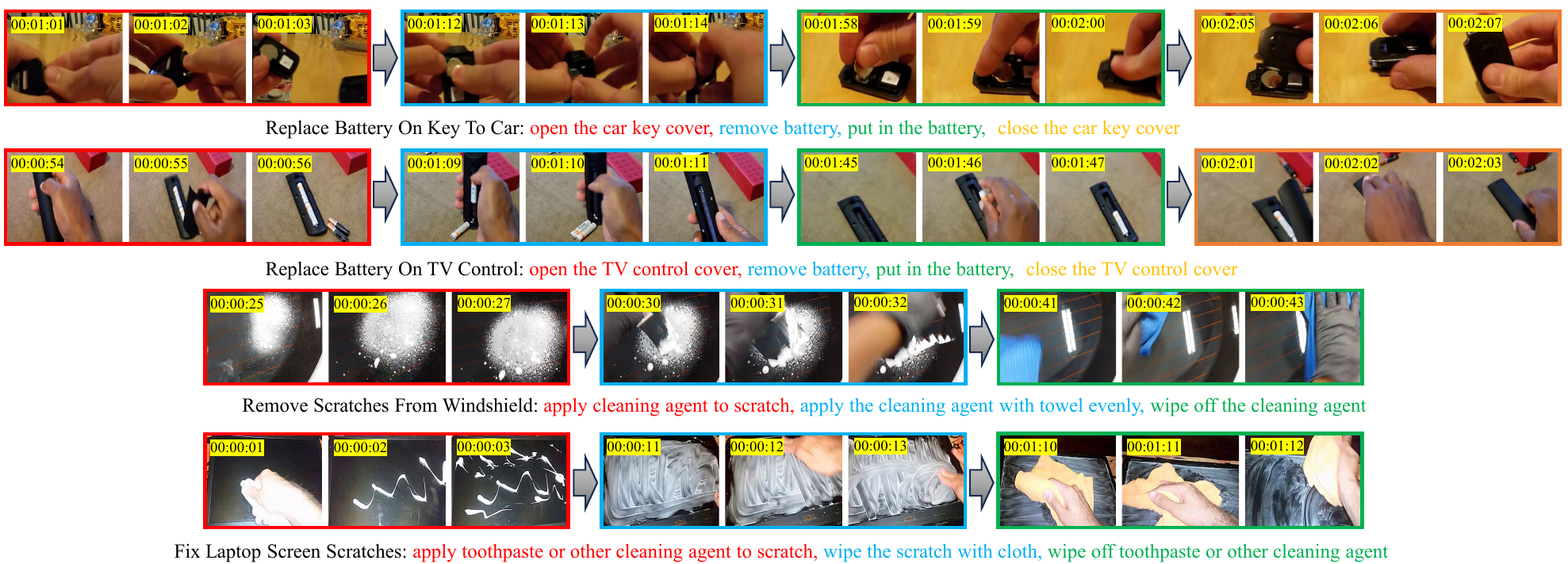}
   \caption{Visualization of OpenEvent. We show examples of two clusters, with every two rows coming from the same cluster. Different actions are marked with different colors.}
   \label{fig:dataset}
   \vspace{-1.2em}
\end{figure}

After the above four stages, we get OpenEvent, which includes 43 events across 8 domains (shown in Table~\ref{tab:datasets}). In general, our OpenEvent contains rich action data covering multiple domains and events. Meanwhile we also ensure the transfer capabilities between similar events, which has not been considered before. More detailed statistics are shown in supplementary materials. We show the visualization results of our dataset in Figure~\ref{fig:dataset}, including videos of 4 events in two clusters. The domains of events in the last two rows are different, which shows that our dataset also attempts to include cross-domain knowledge transfer.

\subsection{Benchmark: evaluation metrics}
\label{sec_metrics}
We follow procedure planning and evaluate the performance using three increasingly strict metrics. (i) mean Intersection over Union (mIoU), which is the least restrictive metric. This metric requires the model to output the right actions without requiring the actions to be in the correct order. We calculate the IoU by $\frac{\left | \left \{ \tilde{a} _i  \right \} \bigcap \left \{ a_i \right \} \right |  }{\left | \left \{ \tilde{a} _i  \right \} \bigcup  \left \{ a_i \right \} \right | }$ on a single sample, where $\left \{ a_i \right \} $ is the set of ground truth actions, and $\left \{ \tilde{a} _i \right \} $ is the set of predicted actions. (ii) Accuarcy (Acc), which considers the accuracy of action at each step. This metric only focuses on the accuracy of each step, and does not require the entire sequence completely correct. (iii) Success Rate (SR), which requires that the entire action sequence is completely consistent with the ground truth. SR is the strictest metric and the most important metric for evaluating procedure planning. Since the testing set of OEPP includes two parts of samples, base and novel, we will calculate three metrics under each part when evaluating the model.

\section{Method}

We deal with the open-event procedure planning problem by considering it as visual-text pair matching task. In this section, we introduce our customized framework for open-event procedure planning in detail. Our framework aims to provide a simple and general baseline without any complex design for the empirical study on this new OEPP task.

\subsection{Overview}
\begin{figure*}[t]
  \centering
   \includegraphics[width=0.7\linewidth]{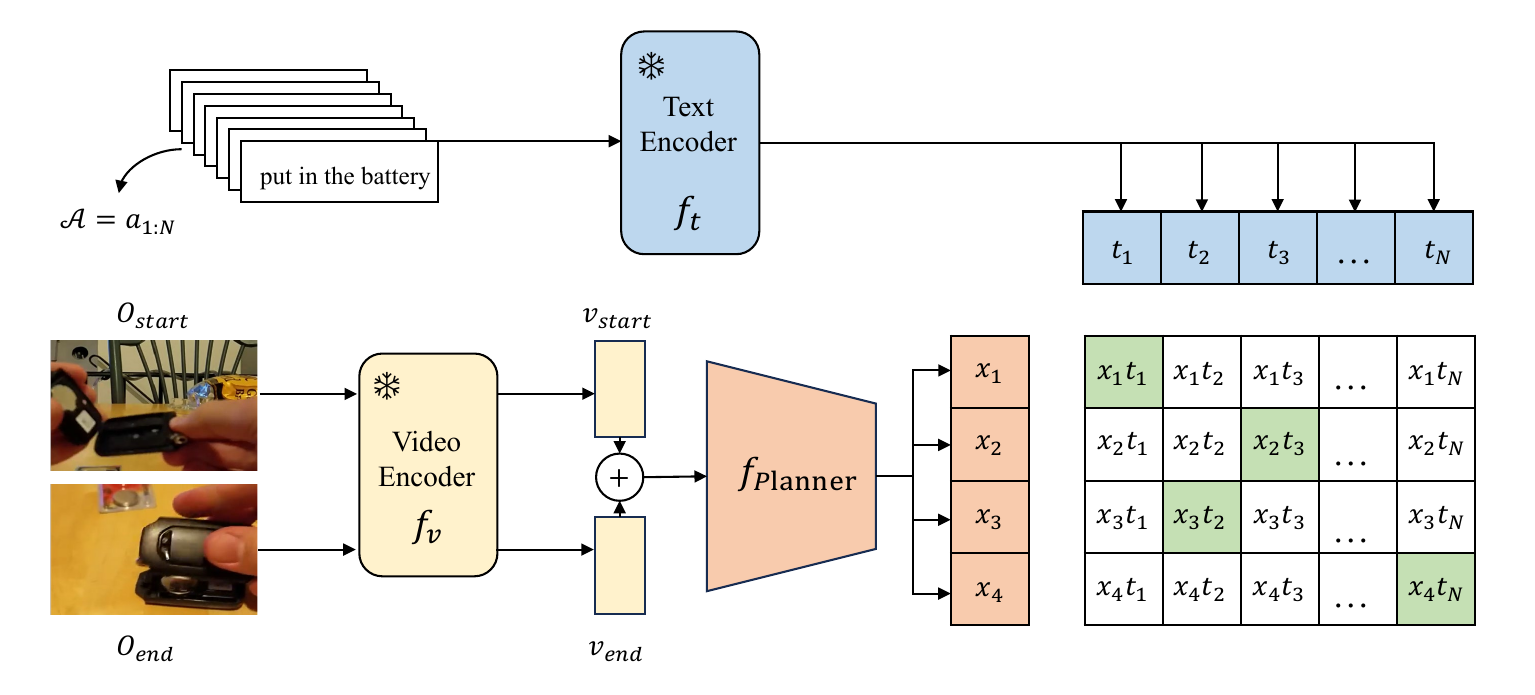}
   \caption{The overview of our framework. When prediction horizon $T=4$, given the start and end observations and the action space, we feed them into video and text encoder separately. Then we use several procedure planners to generate $T$ embeddings and calculate the similarity matrix with the action text features. The green grid in the matrix is the ground truth. }
   \label{fig:otpp}
   \vspace{-10pt}
\end{figure*}

We treat open-event procedure planning as a visual-text pair matching problem just like CLIP~\cite{CILP}, which learns perception from the supervision contained with large-scale image-language pairs. As shown in Figure~\ref{fig:otpp}, for planning horizon $T$, given the action space $\mathcal{A}={a_1,a_2,...,a_N}$ that can be used for planning, we will get $T \times N$ visual-text pairs. Unlike CLIP and other VLMs, we cannot directly obtain the state information of $T$ actions since the input of OEPP only contains the start and end observations($o_{start}$ and $o_{end}$). The intermediate state information is unavailable, so we need to generate the intermediate state information ${x_1,x_2,...,x_T}$ through a planning model and then match the generated results with text. 

\subsection{Video and text encoders}
Open-event procedure planning consumes visual observations {$o_{start}$, $o_{end}$} and action space $\mathcal{A}$ as input. We encode them separately with video and text encoders (frozen) $f_v,f_t$, which makes no assumptions about the encoder architecture and can be used with any video and text backbone. We feed $o_{start}$ and $o_{end}$ into video encoder $f_v$ to obtain video tokens $v_{start}$ and $v_{end}$. Similarly, vectors $\left \{t_1,t_2,...t_N\right \}$ for actions in $\mathcal{A}$ are obtained via text encoder $f_t$. The video tokens and action vectors have the same embedding dimension.
\begin{equation}
    v_{start} = f_v(o_{start}), v_{end} = f_v(o_{end}), o_{start}, o_{end} \in \mathbb{R}^d
\end{equation}

\begin{equation}
    \left \{t_1,t_2,...t_N\right \} = f_t(\left \{a_1,a_2,...a_N\right \}),t_i \in \mathbb{R}^d
\end{equation}

\subsection{Visual and text pair matching}
Given the start and end video tokens $v_{start}, v_{end}$, we apply procedure planning model $f_{Planner}$ to generate visual embedding of $T$ steps, ${x_1,x_2,...,x_T}$. The dimension of $x_i$ is the same as text tokens.
\begin{equation}
    \left \{x_1,x_2,...,x_T\right \} = f_{Planner}(\left \{v_{start},v_{end}\right \}),x_i \in \mathbb{R}^d
\end{equation}

To train this model, we first construct a ground truth matrix $G$ with dimensions $T \times N$ for each sample. The row in the matrix represents the label of the action in the sequence in the form of one-hot encoding. $G_{ij} = 1$ represents that the $i-th$ step is the $j-th$ action in the current action space. Then we calculate the cosine similarity between $T$ visual embeddings and $N$ action text tokens to obtain a prediction matrix $S$, in which the rows are the logits of the action steps. 
\begin{equation}
    S_{ij} = Cosine\_similarity(x_i,t_j), 0<i<T,0<j<N
\end{equation}
After that, we pass the similarity matrix $S$ through Softmax to get the action probability at each step $P$. We calculate the Cross Entropy loss for each row of the prediction matrix $P$ and the ground truth matrix $G$ to obtain the loss. Considering that OEPP is not a complete classification task and we need to generate $T$ embeddings, we also use Mean Squared Error loss to measure the difference between the predicted embedding $x_i$ and the target embedding $t_j$. The final loss is the weighted sum of these two parts, $L_{sum} = \theta_{1} L_{ce} + \theta_{2} L_{mse}, $ where
\begin{equation}
    L_{ce} = \sum_{i=1}^{T}CE\_loss(P_i,G_i), \quad L_{mse} = \sum_{i=1}^{T}MSE\_loss(x_i,t_j).
\end{equation}

\subsection{Implementation details}
\label{details}
We choose the video and text encoder of VideoCLIP~\cite{videoclip} as our feature encoders, $f_v,f_t$, which is pre-trained on HowTo100M~\cite{howto100m}. We train our model with base action space and optimize it for 200 epochs with ADAM~\cite{adam} on a single TITAN XP GPU for each prediction horizon $T$. We then select the best performed model on the validation set as our final model. We conduct inference with the base and novel action spaces, respectively. For each predicted embedding, we select the action with the highest cosine similarity to it. Thus we can still plan a sequence based on the similarity though there are many unseen actions in the novel action space.

\begin{table*}[t]
\caption{Results of Open-event Procedure Planning on OpenEvent for prediction horizon $T=3,4$. Model marked with * means that we reimplement the model based on the open-event settings. }
\centering
\resizebox{0.65\linewidth}{!}{
\begin{tabular}{cccccccc}
\hline
                &                                & \multicolumn{3}{c}{Base}                                                                                      & \multicolumn{3}{c}{Novel}                                                                                      \\ \cline{3-8}
Models & T & SR & Acc & mIoU & SR & Acc & mIoU \\ \hline
Random          & 3          &\textless 0.01                   & 0.91                   &1.26                     &\textless 0.01                      & 0.83                    &1.28                      \\
Matching        & 3          &0.09                   &10.46                   &10.65                     & 0.35                   & 20.97                   & 24.14                     \\
MLP-based           & 3          &         29.44         &       56.09          &   60.43         &        \textbf{11.41}          &   36.45                &       42.42             \\
Transformer-based          & 3          &        26.27         &        55.30            &        59.41          &         \textbf{11.41}          &    \textbf{37.28}                 &       \textbf{43.38}               \\
PDPP*~\cite{PDPP}          & 3          &\textbf{30.76}                  &\textbf{57.09}                    &\textbf{61.90}                     &9.76                   &33.81                    & 37.99                     \\ \hline
Random          & 4          &\textless 0.01                   & 0.84                   & 1.62                    &\textless 0.01                    & 0.77                    & 1.44                     \\
Matching        & 4          &\textless 0.01                   &9.52                   &10.26                     &0.23                    &19.04                     &23.92                      \\
MLP-based            & 4          &          17.79         &        50.42           &     59.98           &       7.48           &  35.06              &          \textbf{43.93}           \\
Transformer-based          & 4         &        15.99          &        49.24           &      57.98          &      \textbf{7.94}          &    \textbf{35.12}                 &       43.88          \\
PDPP*~\cite{PDPP}            & 4          & \textbf{19.48}                  & \textbf{50.96}                   &\textbf{61.64}  & 7.63                  & 32.52                    &  39.71                  \\ \hline
\end{tabular}}
\label{tab1}
\vspace{-10pt}
\end{table*}

\section{Experiments}

In this section, we detail the video curation process and evaluate several methods with our framework on OpenEvent, presenting our experimental results and ablation studies.

\subsection{Video curation}

We follow previous work~\cite {DDN,GAIL,PDPP} and treat each video as image sequence $I_{1:L}$ including $M$ action clips with action labels $a_{1:M}$ and temporal boundaries $\left (  {ts}_{1:M},{te}_{1:M}\right ) $. For the $i$-th action clip, we choose images around the beginning $I_{ts_i:ts_{i+\delta }}$ as $os_i$, and $I_{te_{i-\delta }:te_i}$ as $oe_i$. For procedure planning, we need to select an action sequence of length $T$, but the number of actions in most videos is not equal to $T$. For videos with more than $T$ actions, we curate the videos with a sliding window of time horizon $T$ to consider all procedure plans. For videos whose actions are less than $T$, we repeat the action until the length is equal to $T$. Each sample will thus contain $T$ actions. Among these actions, $os_1$ of the first action is $o_{start}$, and $oe_T$ of the last action is $o_{end}$. Then $o_{start}$ and $o_{end}$ will be used as the input of our model.

\subsection{Quantitative results}

We evaluate several procedure planners with our proposed framework for Open-event procedure planning:

\noindent\textbf{Random}. This method just randomly selects the action sequence from the given action space. The Random baseline shows a lower limit of model performance.

\noindent\textbf{Matching}. The matching baseline directly match the visual and text features obtained from the pre-trained VLMs without any training. Specifically, we calculate the similarity between the visual features of input and the text features of all actions in the action space separately, and select the two actions with the highest similarity score as start and end actions. For the intermediate actions, we first average pool the visual features of the start and end and then calculate the similarity. $T-2$ actions with the highest similarity score will be selected as the intermediate actions.

\noindent\textbf{MLP-based Method}. We use a simple three-layer MLP with $T$ linear heads to predict $T$ embeddings.

\noindent\textbf{Transformer-based Method}. Following previous work~\cite{p3iv,plate,schema}, we construct a transformer-based baseline method, using transformer encoder module and position embedding to enhancing the temporal modeling ability.

\noindent\textbf{PDPP}. For the initial procedure planning task, PDPP aims to fit the distribution of the intermediate action sequence $[a_1,a_2,...,a_T]$, which depends on the given observations, $o_s,o_g$, and task class predicted in advance. PDPP concatenates them along the action feature dimension and thus the model input of PDPP for training can be represented as a multi-dimension array. Each column in the input represents the condition information (task class), action one-hot vector, and corresponding observation for a certain action, as shown in Eq \ref{PDPP_input}.
\begin{equation}
    \begin{bmatrix}
 c &c  &  &c  &c \\
 a_1 &a_2  &...  &a_{T-1}  &a_T \\
  o_s&0  &  &0  &o_g
  \label{PDPP_input}
\end{bmatrix}
\end{equation}

For open-event procedure planning, we modify the input of PDPP in the following two aspects. (i) For open setting, actions can not be represented as one-hot vectors since there are novel actions unseen during inference. Thus we replace the one-hot vectors $[a_1,a_2,...,a_T]$ with the text embedding features of the corresponding step name $[t_1,t_2,...,t_T]$, which is extracted by the pretrained text encoder. (ii) Open-event procedure planning does not provide event names for novel tasks, thus the application of task information in PDPP is not available under the new setting. To deal with this conflict, we delete the task related condition information in PDPP. Note that the ``task'' mentioned in PDPP is the ``event'' here. So the new input of PDPP for open-event procedure planning can be represented as Eq \ref{PDPP_input_}.
\begin{equation}
    \begin{bmatrix}
 t_1 &t_2  &...  &t_{T-1}  &t_T \\
  o_s&0  &  &0  &o_g
  \label{PDPP_input_}
\end{bmatrix}
\end{equation}

For training, in order to maintain a consistence training strategy, we utilize both the $L_{MSE}$ and $L_{CE}$ to train PDPP. For inference, the initial PDPP take out the $[a_1,a_2,...,a_T]$ and select the index of every maximum value, while we take out the $[t_1,t_2,...,t_T]$ to calculate the cosine similarity with the action text embedding in the given action space, and select the action of every maximum similarity.

\noindent\textbf{GPT-4o}. We also use GPT-4o~\cite{gpt-4o} to test on both base and novel test sets. To obtain quantitative metrics, we provided GPT-4o with an optional action space and informed GPT-4o of the number of actions included in the sequence. Given that we do not know the training data of GPT-4o, which has probably seen most of the tasks and actions in the novel test set, comparing GPT-4o with the above baselines is not entirely fair.  We thus include the detailed experimental results of GPT-4o in the supplementary materials.

\begin{table}[t]
\caption{The results of Success Rate on OpenEvent for longer prediction horizon $T$. }
\centering
\resizebox{0.6\linewidth}{!}{
\begin{tabular}{lccccc}
\hline
              &  & T=3            & T=4            & T=5            & T=6           \\ \cline{3-6} 
Models    &Action Space       & SR             & SR             & SR             & SR            \\ \hline
MLP-based     &Base        &  29.44         &    17.79       &   16.12      &  15.68      \\
Transformer-based     &Base        &  26.27         &    15.99     &      15.29      &   14.24       \\
PDPP*~\cite{PDPP} &Base  & \textbf{30.76}  &\textbf{19.48} &\textbf{16.80}  &\textbf{16.48} \\ \hline
MLP-based &Novel  &   \textbf{11.41}        &   7.48       &  5.11       &   6.20        \\
Transformer-based     &Novel        &  \textbf{11.41}         &    \textbf{7.94}       &  6.31     &   5.66       \\
PDPP*~\cite{PDPP} &Novel &9.76  &7.63  &\textbf{6.96}  &\textbf{7.16}  \\ \hline
\end{tabular}}
\label{tab2}

\end{table}

In Table~\ref{tab1}, we evaluate the performance of various baselines for OEPP. It is clear that methods involving training significantly outperform the direct matching baseline, which does not utilize training. PDPP* demonstrates impressive performance under the base testing set and substantially surpasses other methods. However, for novel testing set, PDPP* can not completely outperform other methods. We assume the reason for this is that PDPP* has a strong ability to fit the training distribution, thus its performance is better on base testing set. However, the great fitting ability can limit the knowledge transferable ability of PDPP*, so other methods can perform better under the novel action space, especially for the Acc metric. We also evaluate these methods with longer prediction horizons in Table~\ref{tab2}. It can be seen that PDPP* achieves the best results when $T=5,6$. This is because the design of action sequence modeling in PDPP has greater advantages under longer prediction steps. Overall, the metrics for the novel test set are significantly worse than those for the base test set, indicating that these methods fail to effectively learn knowledge transfer between different events. This also reveals the need to improve procedural knowledge transfer ability for OEPP models.

\subsection{Ablation}
\label{sec_ablation}
\begin{table}[t]
\caption{Ablation study on loss functions for $T=3$ using Transformer-based method}
\centering
\resizebox{0.7\linewidth}{!}{
\begin{tabular}{ccccccccccc}
\hline
       &          & \multicolumn{3}{c}{Val}         & \multicolumn{3}{c}{Base}                                                                               & \multicolumn{3}{c}{Novel}                                                                     \\ \cline{3-11} 
$L_{ce}$ & $L_{mse}$ & \multicolumn{1}{c}{SR} & \multicolumn{1}{c}{Acc} & \multicolumn{1}{c}{mIoU} & \multicolumn{1}{c}{SR} & \multicolumn{1}{c}{Acc} & \multicolumn{1}{c}{mIoU} & \multicolumn{1}{c}{SR} & \multicolumn{1}{c}{Acc} & \multicolumn{1}{c}{mIoU}\\ \hline
    \checkmark   &     & 19.48                            & 50.74                         & 55.70           & 22.32                            & 51.00                          & 54.78                            & 8.69                            & 32.41                            & 38.90                    \\
       &\checkmark         & 23.22                            & 53.68                          & 56.97       & 23.20                           & 53.02                            & 56.39                            & 8.75                  & 34.67                            & 39.70                    \\
 \checkmark      &   \checkmark       & \textbf{25.71}                            & \textbf{55.53}                          & \textbf{59.73}    & \textbf{26.27}          &        \textbf{55.30}            &       \textbf{ 59.41  }         &         \textbf{11.41}          &    \textbf{37.28}                 &       \textbf{43.38}           \\ \hline
\end{tabular}}
\label{ablation}
\end{table}

\noindent\textbf{Loss Function}. In Table~\ref{ablation}, we investigate the effect of loss function, where each of the following losses is adopted: $L_{mse}$ only, $L_{ce}$ only and the combination of the two losses. We found that using the weighted sum of $L_{mse}$ and $L_{ce}$ best facilitates the model for both base and novel events. Using $L_{mse}$ only results in better performance than using $L_{ce}$ only, considering that the target of $L_{mse}$ is to fit the action feature in the pre-trained VLMs space, which is more suitable for the OEPP. Based on the experiment results, our final loss function is designed as the weighted sum of $L_{mse}$ and $L_{ce}$.

\begin{table}[t]
\caption{Ablation study on features for $T=3$}
\centering
\resizebox{0.9\linewidth}{!}{
\begin{tabular}{ccccccccccc}
\hline
       &       & \multicolumn{3}{c}{Val}         & \multicolumn{3}{c}{Base}                                                                               & \multicolumn{3}{c}{Novel}                                                                     \\ \cline{3-11} 
Method & Features & \multicolumn{1}{c}{SR} & \multicolumn{1}{c}{Acc} & \multicolumn{1}{c}{mIoU} & \multicolumn{1}{c}{SR} & \multicolumn{1}{c}{Acc} & \multicolumn{1}{c}{mIoU} & \multicolumn{1}{c}{SR} & \multicolumn{1}{c}{Acc} & \multicolumn{1}{c}{mIoU} \\ \hline
    Transformer-based   &  MIL-NCE \cite{MIL-NCE}     & 14.50                            & 43.53                            & 48.82         & 16.26                            & 44.93                            & 48.54                             & 3.31                            & 25.74                             & 31.76                    \\
       Transformer-based   &  VideoCLIP \cite{videoclip}   & \textbf{25.71}                            & \textbf{55.53}                          & \textbf{59.73}           & \textbf{26.27}          &        \textbf{55.30}            &       \textbf{59.41}         &         \textbf{11.41}          &    \textbf{37.28}                 &       \textbf{43.38}                   \\ \hline
PDPP* \cite{PDPP}  &  MIL-NCE \cite{MIL-NCE}          &19.37	&47.49	&52.64 &  17.93	&45.31	&49.93	&5.32	&25.67	&31.57	
                               \\
       PDPP* \cite{PDPP}   &  VideoCLIP \cite{videoclip}       & \textbf{30.92}                            & \textbf{56.44}                          & \textbf{61.25}       &\textbf{30.76}                  &\textbf{57.09}                    &\textbf{61.90}                     &\textbf{9.76}                   &\textbf{33.81}                    & \textbf{37.99}                    \\ \hline
\end{tabular}}
\label{feature}
\end{table}

\noindent\textbf{Feature Encoders}. In Table~\ref{feature}, we conduct ablation experiments with different feature encoders, opting for MIL-NCE~\cite{MIL-NCE}, a VLM that has also been pretrained on the Howto100M~\cite{howto100m} dataset. The video encoder used is S3D~\cite{s3d}, and the text encoder is Word2Vec~\cite{word}. The experiments demonstrate that VideoCLIP better aligns visual and textual information, yielding superior results for OEPP.

\begin{figure}[t]
 \centering
  \includegraphics[width=0.7\linewidth]{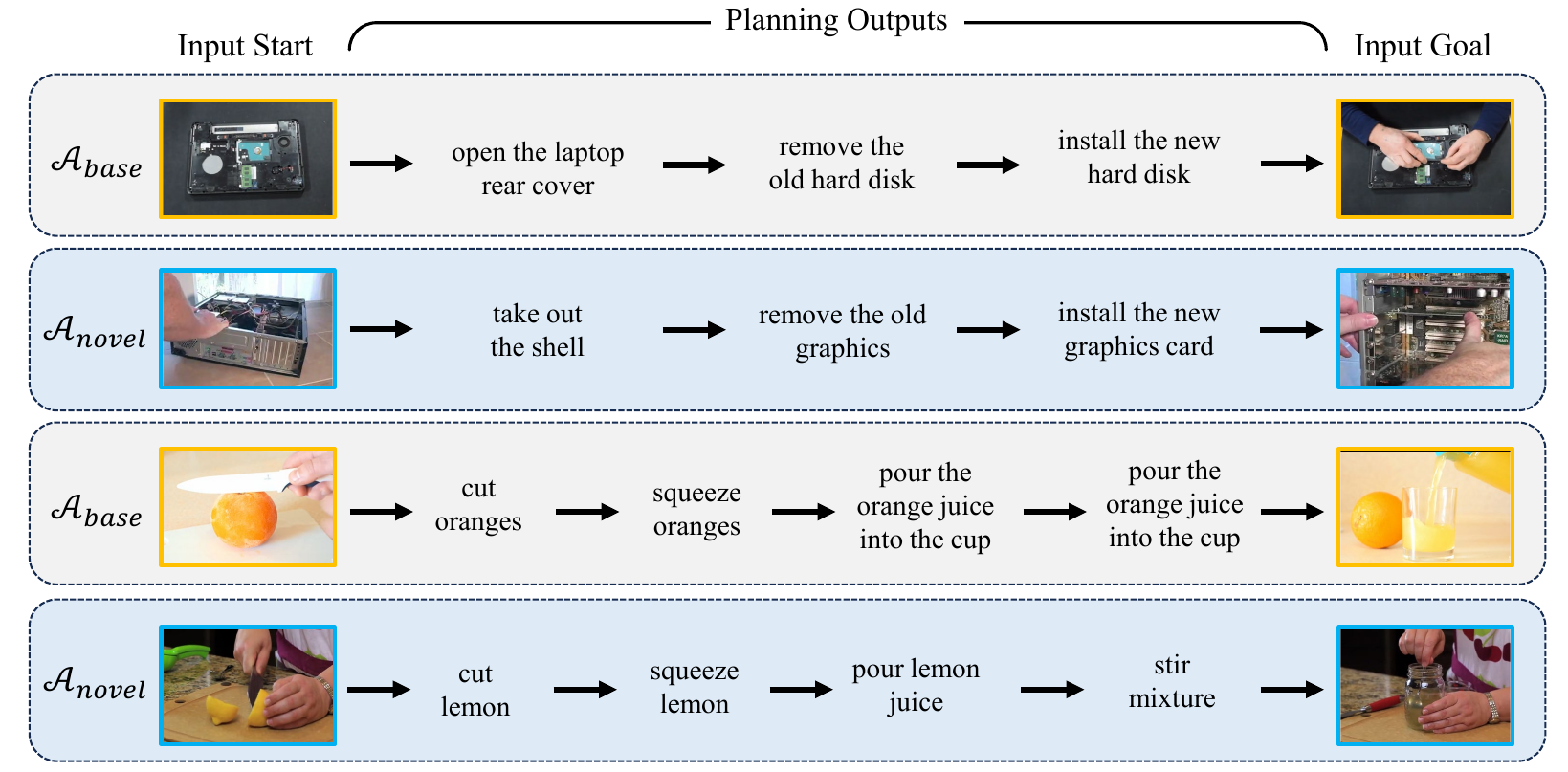}
  \caption{Visualization of successful results on OpenEvent for $T=3,4$.}
  \label{fig:vis}
\end{figure}

\subsection{Qualitative results}

In Figure~\ref{fig:vis}, we present the visualizations of our planning results on OpenEvent for prediction horizons $T=3,4$. For each prediction horizon, we select one sample from the base and novel events respectively, and in order to reflect that our method has learned procedural knowledge, we select two samples from the same cluster.

\section{Limitation and conclusion}
In this section, we discuss the limitations of our work and summarize our conclusion. The first limitation is that our dataset is based on existing datasets, which means that we have a limited number of events. To address the problem of procedure planning in the open-event setting more effectively, we wish to expand our dataset further and explore the utilization of instructive articles from platforms like WikiHow in order to provide a large-scale procedural knowledge base. Additionally, we must acknowledge that our task is still an understanding task rather than a generative task, which means the novel action space is still needed for planning for unseen tasks. We hope that in the future, open-event procedure planning can be extended to a generative task containing a wider range of events and domains, thus be more general and practical.

In this paper, we propose a new task termed as Open-event Procedure Planning (OEPP), which extends the traditional procedure planning to an open-event setting. To better promote the progress of OEPP, we rebuild a new instructional video benchmark OpenEvent using existing datasets. We carefully design the principles to build a reasonable benchmark by ensuring the transfer possibility between events. Based on our experimental results, we find that it is possible for the model to understand the essential procedural knowledge in an open-event setting. We believe that with the development of VLMs and LLMs, the open-event setting can become increasingly meaningful and practical in the future.

\bibliographystyle{unsrt}
\bibliography{neurips_data_2024}

\clearpage
\section{Appendix}

\subsection{Data and code}

Our data and code: https://github.com/FOXamber/OEPP

\subsection{Experiment}
\label{exp}

\subsubsection{Implementation details}

Our loss function is $L_{sum} = \theta_{1} L_{ce} + \theta_{2} L_{mse}.$ During training, we set the two parameters to 1.0 and 0.2 for three baselines. The learning rate is set to 0.0001 for MLP-based and Transformer-based methods, and to 0.0005 for PDPP*. For more details, please refer to the code.

\subsubsection{More results for different horizons}

\begin{table}[h]
\caption{Results of Open-event Procedure Planning on OpenEvent for prediction horizon $T \in \{3,4,5,6\}$. The model marked with an asterisk (*) means that we reimplement the model based on the open-event settings. }
\centering
\resizebox{0.7\linewidth}{!}{
\begin{tabular}{cccccccc}
\hline
                &                                & \multicolumn{3}{c}{Base}                                                                                      & \multicolumn{3}{c}{Novel}                                                                                      \\ \cline{3-8}
Models & T & SR & Acc & mIoU & SR & Acc & mIoU \\ \hline
MLP-based           & 3          &         29.44         &       56.09          &   60.43         &        \textbf{11.41}          &   36.45                &       42.42             \\
Transformer-based          & 3          &        26.27         &        55.30            &        59.41          &         \textbf{11.41}          &    \textbf{37.28}                 &       \textbf{43.38}               \\
PDPP*~\cite{PDPP}          & 3          &\textbf{30.76}                  &\textbf{57.09}                    &\textbf{61.90}                     &9.76                   &33.81                    & 37.99                     \\ \hline
MLP-based            & 4          &          17.79         &        50.42           &     59.98           &       7.48           &  35.06              &          \textbf{43.93}           \\
Transformer-based          & 4         &        15.99          &        49.24           &      57.98          &      \textbf{7.94}          &    \textbf{35.12}                 &       43.88          \\
PDPP*~\cite{PDPP}            & 4          & \textbf{19.48}                  & \textbf{50.96}                   &\textbf{61.64}  & 7.63                  & 32.52                    &  39.71                  \\ \hline
MLP-based            & 5          &          16.12         &        48.54           &       63.48           &        5.11           &     35.15               &        43.44           \\
Transformer-based          & 5         &        15.29          &        48.87           &      62.05          &      6.31          &    \textbf{36.10}                 &       \textbf{45.89}              \\
PDPP*~\cite{PDPP}             & 5          & \textbf{16.80}                  & \textbf{49.94}                   &\textbf{65.32}  & \textbf{6.96}                  & 32.76                    &  39.62                  \\ \hline
MLP-based            & 6          &    15.68         &     49.23           &     \textbf{64.96}           &     6.20           &     35.10               &         43.95          \\
Transformer-based          & 6         &        14.24          &        47.65          &      62.40          &      5.66          &    \textbf{35.36}                 &       \textbf{45.41}              \\
PDPP*~\cite{PDPP}             & 6          & \textbf{16.48}                  & \textbf{49.44}                   &63.88  & \textbf{7.16}                  & 33.76                    &  41.33                  \\ \hline
\end{tabular}}
\label{tab:s1}
\end{table}

In Table~\ref{tab:s1}, we show the entire results of our baselines for prediction horizon $T=3,4,5,6$.

\subsubsection{Error bar}
We test the error bars of different models using different random seeds. For T=3, the error bar of the MLP-based method is slightly larger compared to Transformer-based method. The error for Transformer-based method on SR is no more than 1\%, while for MLP-based method it is around 2\%.

\subsubsection{Results of GPT}

\begin{table*}[h]
\caption{Results of GPT-4o for OEPP. The first row presents the results without sliding window.}
\centering
\resizebox{0.75\linewidth}{!}{
\begin{tabular}{ccccccccc}
\hline
                &                    &            & \multicolumn{3}{c}{Base}                                                                                      & \multicolumn{3}{c}{Novel}                                                                                      \\ \cline{4-9}
Models & T &Num of images & SR & Acc & mIoU & SR & Acc & mIoU \\ \hline
GPT-4o~\cite{gpt-4o}          & -     &1     &8.05                  & 21.70                   &37.91                     &13.93                      & 30.38                    &49.52                      \\ \hline
GPT-4o~\cite{gpt-4o}          & 3     &1     &4.27                  & 24.64                   &32.23                     &9.11                      & 34.12                    &49.46                      \\
GPT-4o~\cite{gpt-4o}          & 3     &3     &3.63                  & 22.72                   &32.70                     &7.45                      & 31.87                    &48.02                      \\ \hline
GPT-4o~\cite{gpt-4o}          & 4     &1     &1.40                  & 18.86                   &30.56                     &3.93                      & 29.30                    &54.31                      \\
GPT-4o~\cite{gpt-4o}            & 4     &3     & 0.82                  & 18.74                   &31.92  & 3.24                  &27.04                    &  50.56                  \\ \hline
\end{tabular}}
\label{tab:gpt}
\end{table*}

We use GPT-4o~\cite{gpt-4o} to test on both base and novel test sets. To obtain quantitative metrics, we provided GPT-4o with an optional action space and informed GPT-4o of the number of actions included in the sequence. Given that we do not know the training data of GPT-4o, which has probably seen most of the tasks and actions in the novel test set, comparing GPT-4o with the above baselines is not entirely fair. We report the evaluation results with and without the using of sliding window data curation. We also ablate the results by selecting 1 or 3 frames from the start and end video clips as input observations. The prompt is as followed:

\begin{quote}
\textbf{System:}

You are a helpful assistant, an expert in answering questions about action planning in instructional videos. Based on the start and end images, you need to infer the actions to transform from the start images to the end images. You must choose from the following actions [action\_pool]. Output the name of the actions step by step. 

Example: 1. cut in half 2. slice the pulp

Example: 1. dip detergent with rag or apply detergent 2. clean the floor 3. wash the floor

\textbf{User:}

Infer the actions to transform from the start images to the end images. The actions must be actions from the given action pool.

The number of actions is [T]. The start images: [images encoded with base64]. The end images: [images encoded with base64].

\textbf{Assistant:}

1. open the car key cover 2. take out the car key battery 3. put in the battery 4. close the car key cover

\end{quote}

As shown in Table \ref{tab:gpt}, the planning results without using the sliding window data curation are better. We believe this is because GPT's pre-training samples typically consist of complete action sequences, so using sliding windows to extract partial sequences might degrade its performance. Additionally, the results on novel set are much better. We attribute this to two reasons. First, since we conduct zero-shot evaluation with GPT-4o, there is no difference between planning on the base or novel sets and GPT has likely seen all these actions during training with a high probability. Therefore, the results of the novel set will not be very poor. Second, we request GPT-4o to select actions from the given action space, and the action space in the novel set contains far fewer actions than in the base set (base: 122; novel: 55), making planning in the novel set much easier. Additionally, we find that when more images are provided as observation, the planning results get worse. This reflects that for planning problems, GPT-4o can not better understand the meaning of multiple image sequences than single image. Overall, the performance of GPT on OEPP is not ideal. This also reveals the need to improve procedural knowledge transfer ability for OEPP models.

\subsection{Dataset}

\label{dataset}

\subsubsection{More visualization of OpenEvent}

\begin{figure*}[h]
  \centering
   \includegraphics[width=1\linewidth]{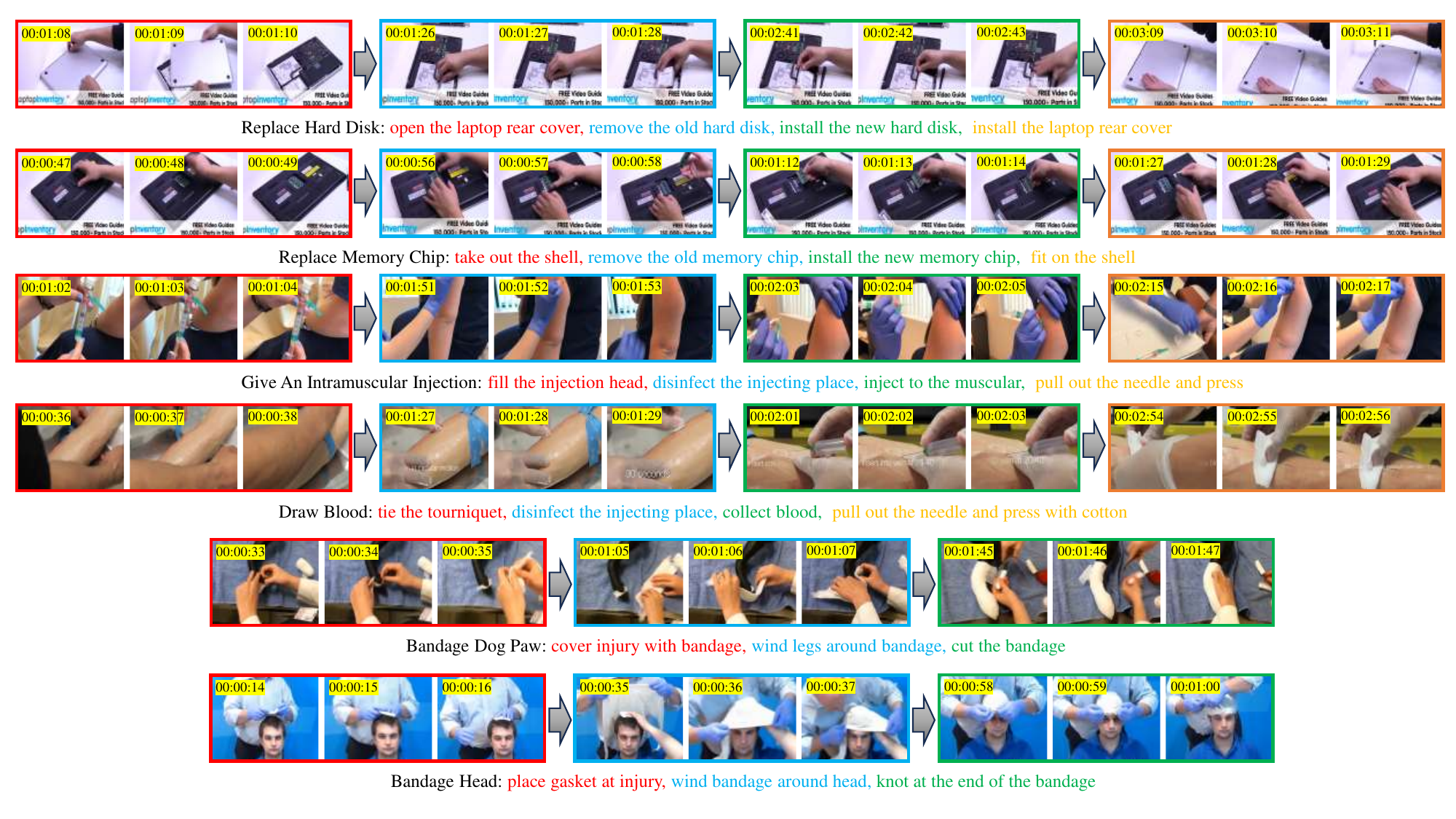}
   \caption{Visualization of OpenEvent.}
   \label{fig:vis}
\end{figure*}

In Figure \ref{fig:vis}, we present more visualization results of OpenEvent. We show more examples of three clusters, with every two rows coming from the same cluster. We show some frames sampled from a video, with each row representing a video. Specifically, we sample one frame for each action in each video for three consecutive seconds, and different actions are marked with different colors.

\subsubsection{The algorithm of text similarity clustering}
In our main paper, we construct our dataset OpenEvent with four stage, the first stage we divide different events into different clusters based on the text similarity, the Alg~\ref{alg1} shows the clustering algorithm.

\begin{algorithm}[t]
    \caption{Text similarity clustering algorithm}
    \label{alg1}
    \begin{algorithmic}[1]
    \REQUIRE $Sentences$: a set of N sentences,$\theta$
    \ENSURE $Clusters$ : a list of clusters
    \FOR {$sentence \in Sentences$}
    \STATE $is\_similar \gets False$
    \FOR {$cluster \in Clusters$}
    \STATE $sim \gets avg\_text\_sim(sentence, cluster)$
    \IF {$sim > \theta$ }
    \STATE $is\_similar \gets True$
    \STATE $cluster.append(sentence)$
    \STATE $\textbf{break}$
    \ENDIF    
    \ENDFOR
    \IF {$is\_similar = False$}
    \STATE $new\_cluster \gets []$
    \STATE $new\_cluster.append(sentence)$
    \STATE $Clusters.append(new\_cluster)$
    \ENDIF
    \ENDFOR
    \end{algorithmic}
    \label{alg1}
\end{algorithm}

\begin{table}[t]
\caption{Statistics of OpenEvent Benchmark.}
\centering
\resizebox{0.5\linewidth}{!}{
\begin{tabular}{ccccc}
\hline
\textbf{OpenEvent}  & \textbf{Events} & \textbf{Actions} & \textbf{Videos} & \textbf{Segments} \\ \hline
Total                             &43               &161                & 2771               & 12210                \\ \hline
Train                       & 29              & 122               &  1285             & 5833                \\
Val                       & 29              & 122               &  337             & 1479                \\
Test\_base                 &  29               & 122                 & 416                &1888                   \\
Test\_novel                   & 14                & 55                 &   733              & 3010                  \\ \hline
\end{tabular}}
\label{tab:stat}
\end{table}

\subsubsection{Statistics of OpenEvent}

\begin{figure}[t]
  \centering
   \includegraphics[width=0.9\linewidth]{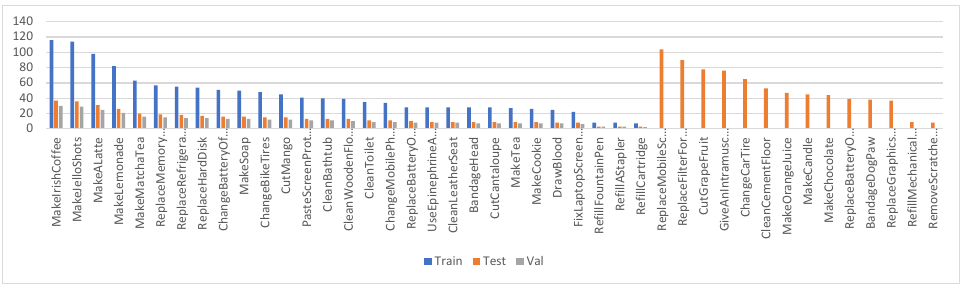}
   \caption{The sample distributions of all the events in OpenEvent.}
   \label{fig:dataset_ov}
\end{figure}

In Table~\ref{tab:stat}, we show the detail statistics of OpenEvent. In Figure~\ref{fig:dataset_ov}, we show the sample distributions of OpenEvent.

\subsubsection{Details of event clusters and action description}
In Table~\ref{cluster}, we present the detail event clusters of our dataset. In Table~\ref{action}, we list all refined actions with the old action labels and the new action labels. 
\begin{table}[h]
\caption{The event clusters of OpenEvent. }
\centering
\resizebox{0.5\linewidth}{!}{
\begin{tabular}{lll}
\hline
\textbf{Domain}     & \textbf{Cluster} & \textbf{Event}                 \\ \hline
Nursing and Care     & 0                & BandageDogPaw                  \\
Nursing and Care     & 0                & BandageHead                    \\
Nursing and Care     & 1                & GiveAnIntramuscularInjection   \\
Nursing and Care     & 1                & UseEpinephrineAuto-injector    \\
Nursing and Care     & 1                & DrawBlood                      \\
Vehicle              & 2                & ChangeCarTire                  \\
Vehicle              & 2                & ChangeBikeTires                \\
Vehicle              & 3                & RemoveScratchesFromWindshield  \\
Gadgets              & 3                & FixLaptopScreenScratches       \\
Gadgets              & 4                & RefillMechanicalPencils        \\
Gadgets              & 4                & RefillFountainPen              \\
Gadgets              & 4                & RefillAStapler                 \\
Electrical Appliance & 4                & RefillCartridge                \\
Gadgets              & 5                & ChangeBatteryOfWatch           \\
Gadgets              & 5                & ReplaceBatteryOnTVControl      \\
Gadgets              & 5                & ReplaceBatteryOnKeyToCar       \\
Gadgets              & 5                & ChangeMobilePhoneBattery       \\
Gadgets              & 6                & ReplaceMobileScreenProtector   \\
Electrical Appliance & 6                & PasteScreenProtectorOnPad      \\
Electrical Appliance & 7                & ReplaceGraphicsCard            \\
Electrical Appliance & 7                & ReplaceMemoryChip              \\
Electrical Appliance & 7                & ReplaceHardDisk                \\
Electrical Appliance & 8                & ReplaceFilterForAirPurifier    \\
Electrical Appliance & 8                & ReplaceRefrigeratorWaterFilter \\
Science and Craft    & 9                & MakeCandle                     \\
Science and Craft    & 9                & MakeSoap                       \\
Pets and Fruit       & 10               & CutMango                       \\
Pets and Fruit       & 10               & CutGrapeFruit                  \\
Pets and Fruit       & 10               & CutCantaloupe                  \\
Drink and Snack      & 11               & MakeMatchaTea                  \\
Drink and Snack      & 11               & MakeTea                        \\
Drink and Snack      & 11               & MakeALatte                     \\
Drink and Snack      & 11               & MakeIrishCoffee                \\
Drink and Snack      & 11               & MakeOrangeJuice                \\
Drink and Snack      & 11               & MakeLemonade                   \\
Drink and Snack      & 11               & MakeJelloShots                 \\
Drink and Snack      & 12               & MakeChocolate                  \\
Drink and Snack      & 12               & MakeCookie                     \\
Housework            & 13               & CleanWoodenFloor               \\
Housework            & 13               & CleanCementFloor               \\
Housework            & 13               & CleanToilet                    \\
Housework            & 13               & CleanBathtub                   \\
Housework            & 13               & CleanLeatherSeat               \\ \hline
\end{tabular}}
\label{cluster}
\end{table}

\begin{table}[t]
\caption{The action description refinement of OpenEvent. }
\centering
\resizebox{0.9\linewidth}{!}{
\begin{tabular}{lll}
\hline
\textbf{Event}                & \textbf{Action Label}                      & \textbf{New Action Label}                   \\ \hline
BandageDogPaw                 & wind legs with bandage                     & wind legs around bandage                    \\
DrawBlood                     & disinfect                                  & disinfect the injecting place               \\
RemoveScratchesFromWindshield & spray the cleaning agent on the car window & apply cleaning agent to scratch             \\
FixLaptopScreenScratches      & wipe the toothpaste                        & wipe off toothpaste or other cleaning agent \\
RefillMechanicalPencils       & remove cap                                 & take off the cap                            \\
RefillMechanicalPencils       & buckle the cap                             & close the cap                               \\
RefillCartridge               & take out the label                         & remove label                                \\
ChangeBatteryOfWatch          & open the back cover                        & open the watch cover                        \\
ReplaceBatteryOnTVControl     & open cover                                 & open the TV control cover                   \\
ReplaceBatteryOnTVControl     & put battery in                             & put in the battery                          \\
ReplaceBatteryOnTVControl     & close cover                                & close the  TV control   cover               \\
ReplaceBatteryOnKeyToCar      & take out the car key battery               & remove battery                              \\
ChangeMobilePhoneBattery      & take down the old battery                  & remove battery                              \\
ChangeMobilePhoneBattery      & load a new battery                         & put in the battery                          \\
PasteScreenProtectorOnPad     & wipe screen                                & wipe the screen                             \\
PasteScreenProtectorOnPad     & wipe screen again                          & wipe the screen                             \\
MakeSoap                      & put the melted soap block into the vessel  & pour the melted soap block into the vessel  \\
CutGrapeFruit                 & remove the peel                            & peel                                        \\
MakeTea                       & prepare and boil water                     & boil water                                  \\
MakeTea                       & prepare and add the tea                    & add tea                                     \\
MakeTea                       & add some water to the tea                  & pour water                                  \\
MakeTea                       & add some ingredients to the tea            & add some ingredients                        \\
MakeOrangeJuice               & juice the oranges                          & squeeze oranges                             \\
MakeCookie                    & pour raw materials                         & add raw materials                           \\
CleanWoodenFloor              & mop the floor                              & clean the floor                             \\
CleanToilet                   & scrub the toilet interior                  & wipe the toilet interior                    \\
CleanBathtub                  & scrub the bathtub                          & wipe the bathtub                            \\
CleanLeatherSeat              & wipe the detergent                         & wipe the leather                            \\ \hline
\end{tabular}}
\label{action}
\end{table}

\subsection{License}

The dataset~\cite{COIN,crosstask} we are using is collected from publicly accessible source. We have followed all legal requirements to integrate this data into our research, emphasizing the importance of transparency in data licensing for proper attribution and appropriate use.

\end{document}